\documentclass[twoside]{article}

\usepackage[accepted]{aistats2021}
\usepackage{amsmath}
\usepackage{amssymb}
\usepackage[T1]{fontenc}
\usepackage[round]{natbib}

\usepackage[utf8]{inputenc} 
\usepackage[T1]{fontenc}    
\usepackage{hyperref}       
\usepackage{url}            
\usepackage{booktabs}       
\usepackage{amsfonts}       
\usepackage{amsmath}
\usepackage{makecell}
\usepackage{graphicx} 
\usepackage{caption,subcaption}
\usepackage{nicefrac}       
\usepackage{microtype}      
\usepackage[inline]{enumitem} 
\usepackage{todonotes}
\usepackage[normalem]{ulem}
\usepackage{multirow}
\makeatletter
\def\maketag@@@#1{\hbox{\m@th\normalfont\normalsize#1}}
\newcommand\Label[1]{&\refstepcounter{equation}(\theequation)\ltx@label{#1}&}
\makeatother

\usepackage{enumitem}

\usepackage{scalerel}

\usepackage{multirow}

\makeatletter

\usepackage{tikz}
\usetikzlibrary{shapes.misc, positioning}
\usetikzlibrary{backgrounds,fit,matrix, calc}
\tikzset{
   box/.style = {minimum height=10pt, minimum width=10pt, draw, rounded corners,rectangle, fill=white!50},
}
\usetikzlibrary{fadings}

\tikzset{
   boxconv/.style = {minimum height=2cm, minimum width=2cm, draw, line width=0.4mm, fill opacity=0.9, rounded corners,rectangle, fill=white!50},
}

\tikzset{
   boxconv_inactive/.style = {minimum height=2cm, minimum width=2cm,line width=0.3mm, draw, line width=0.1mm , fill opacity=0.9, rounded corners,rectangle, gray, fill=white!50},
}
\tikzset{
   input/.style = {minimum height=3cm, minimum width=3cm, draw, , fill opacity=0.9, rectangle, fill=white!50},
}

\tikzset{
   boxpooled/.style = {minimum height=1.5cm, minimum width=1.5cm, draw, line width=0.4mm, fill opacity=0.9, rounded corners,rectangle, fill=white!50},
}

\tikzset{
   boxpooled_inactive/.style = {minimum height=1.5cm, minimum width=1.5cm, draw, line width=0.4mm, fill opacity=0.9, rounded corners,rectangle, gray, fill=white!50},
}

\usetikzlibrary{fadings}\tikzset{
    boxwta/.style={%
        draw=black, thick,
        rectangle,
        rounded corners,
        minimum height=3cm,
        minimum width=3cm
    }
}

\usetikzlibrary{fadings}\tikzset{
    box1/.style={%
        draw=black, thick,
        rectangle,
        minimum height=2cm,
        minimum width=2cm
    }
}

\usetikzlibrary{fadings}\tikzset{
    box2/.style={%
        draw=black, thick,
        rectangle,
        minimum height=1.cm,
        minimum width=1.cm
    }
}

\usetikzlibrary{fadings}\tikzset{
    box3/.style={%
        draw=black, thick,
        rectangle,
        minimum height=.8cm,
        minimum width=.8cm
    }
}
\usetikzlibrary{decorations.markings}

\begin{document}
	
	%
	
	%
	\runningauthor{Konstantinos P. Panousis, Sotirios Chatzis, Antonios Alexos, Sergios Theodoridis}
	
	\twocolumn[
	
	\aistatstitle{Local Competition and Stochasticity for Adversarial Robustness in Deep Learning}
	
	\aistatsauthor{ Konstantinos P. Panousis\textsuperscript{\textdagger} \And Sotirios Chatzis\textsuperscript{\textdagger} \And  Antonios Alexos\textsuperscript{\ddag} \\ \And Sergios Theodoridis\textsuperscript{\textsection} }
	\vspace{.5ex}
	\aistatsaddress{ \textsuperscript{\textdagger}Cyprus University of Technology, Limassol, Cyprus\\                 \textsuperscript{\ddag}University of California Irvine, CA, USA\\
		\textsuperscript{\textsection}National and Kapodistrian University of Athens, Athens, Greece \& Aalborg University, Denmark\\
		\texttt{k.panousis@cut.ac.cy}
	} ]
	
	\begin{abstract}
		This work addresses adversarial robustness in deep learning by considering deep networks with stochastic local winner-takes-all (LWTA) activations. This type of network units result in sparse representations from each model layer, as the units are organized in blocks where only one unit generates a non-zero output. The main operating principle of the introduced units lies on stochastic arguments, as the network performs posterior sampling over competing units to select the winner. We combine these LWTA arguments with tools from the field of Bayesian non-parametrics, specifically the stick-breaking construction of the Indian Buffet Process, to allow for inferring the sub-part of each layer that is essential for modeling the data at hand. Then, inference is performed by means of stochastic variational Bayes. We perform a thorough experimental evaluation of our model using benchmark datasets. As we show, our method achieves high robustness to adversarial perturbations, with state-of-the-art performance in powerful adversarial attack schemes.
	\end{abstract}

	\section{Introduction}
	
	Despite their widespread success, Deep Neural Networks (DNNs) are notorious for being highly susceptible to adversarial attacks.  \textit{Adversarial examples}, i.e. inputs comprising carefully crafted perturbations, are designed with the aim of ``\textit{fooling}'' a considered model into \textit{misclassification}. It has been shown that, even small perturbations in an original input, e.g. via an $\ell_p$ norm, may successfully render a DNN vulnerable; this highlights the frailness of common DNNs \citep{papernot2017practical}. This vulnerability casts serious doubts regarding the confident use of modern DNNs in safety-critical applications, such as autonomous driving \citep{boloor2019simple, chen2015deepdriving}, video recognition \citep{jiang2019black}, healthcare \citep{finlayson2019adversarial} and other real-world scenarios \citep{kurakin2016adversarial}. To address these concerns, significant research effort has been devoted to adversarially-robust DNNs. 
	
	Adversarial attacks, and associated defense strategies, comprise many different approaches sharing the same goal; make deep architectures more \textit{reliable} and \textit{robust}. In general, adversarially-robust models are obtained via the following types of strategies: (i) \textit{Adversarial Training}, where a model is trained with both original and perturbed data \citep{madry2017towards,tramer2017ensemble,shrivastava2017learning}; (ii) \textit{Manifold Projections}, where the original data points are projected onto a different subspace, presuming that, therein, the effects of the perturbations can be mitigated \citep{jalal2017robust, shen2017ape, song2017pixeldefend}; (iii) \textit{Stochastic Modeling}, where some randomization of the input data and/or the neuronal activations is performed on each hidden layer \citep{prakash2018deflecting, dhillon2018stochastic, xie2017mitigating}; and (iv) \textit{Preprocessing}, where some aspects of either the data or the neuronal activations are modified to induce robustness \citep{buckman2018thermometer, guo2017countering, kabilan2018vectordefense}. 
	
	Despite these advances, most of the currently considered approaches and architectures are particularly tailored to the specific characteristics of a considered type of attacks. This implies that such a model may fail completely in case the adversarial attack patterns change in a radical manner. To overcome this challenge, we posit that we need to devise an activation function paradigm different from common neuronal activation functions, especially ReLU.
	
	Recently, the deep learning community has shown fresh interest in more biologically plausible models. In this context, there is an increasing body of evidence from Neuroscience that neurons with similar functions in a biological system are aggregated together in blocks, and local competition takes place therein for their activation (Local-Winner-Takes-All, LWTA, mechanism). Under this scheme, in each block, only one neuron can be active at a given time, while the rest are inhibited to silence. Crucially, it appears that this mechanism is of stochastic nature, in the sense that the same system may produce different neuron activation patterns when presented with exactly the same stimulus at multiple times \citep{kandel2000principles, andersen1969participation, stefanis1969interneuronal, douglas2004neuronal, lansner2009associative}. Previous implementations of the LWTA mechanism in deep learning have shown that the obtained sparse representations of the input are quite informative for classification purposes \citep{lee1999learning, olshausen1996emergence}, while exhibiting \textit{automatic gain control}, \textit{noise suppression} and \textit{robustness to catastrophic forgetting} \citep{srivastava2013compete, grossberg1982contour, carpenter1988art}. However, previous authors have not treated the LWTA mechanism under a systematic stochastic modeling viewpoint, which is a key component in actual biological systems. 
	
	Finally, recent works in the community, e.g. \cite{verma2019error}, have explored the susceptibility of the conventional one-hot output encoding of deep learning classifiers to adversarial attacks. By borrowing arguments from coding theory, the authors examine the effect of encoding the deep learning classifier output using \textit{error-correcting output codes} in the adversarial scenario. The experimental results suggest that such an encoding technique enhances the robustness of a considered architecture, while retaining high classification accuracy in the benign context. 
	
	Drawing upon these insights, in this work we propose a new deep network design scheme that is particularly tailored to address adversarially-robust deep learning. Our approach falls under the stochastic modeling type of approaches. Specifically, we propose a deep network configuration framework employing: (i) stochastic LWTA activations; and (ii) an Indian Buffet process (IBP)-based mechanism for learning which sub-parts of the network are essential for data modeling. We combine this modeling rationale with Error Correcting Output Codes \citep{verma2019error} to further enhance performance.
	
	We evaluate our approach using well-known benchmark datasets and network architectures. We provide related source code at: \url{https://github.com/konpanousis/adversarial_ecoc_lwta}. 
	The obtained empirical evidence vouches for the potency of our approach, yielding state-of-the-art robustness against powerful benchmark attacks. The remainder of the paper is organized as follows: In Section 2, we introduce the necessary theoretical background. In Section 3, we introduce the proposed approach and describe its rationale and inference algorithms. In Section 4, we perform extensive experimental evaluations, providing insights for the behavior of the produced framework. In Section 5, we summarize the contribution of this work.
	
	\section{Technical Background}
	
	\subsection{Indian Buffet Process}
	The Indian Buffet Process (IBP) \citep{ghahramani2006infinite} defines a probability distribution over infinite binary matrices. IBP can be used as a flexible prior for latent factor models, allowing the number of involved latent features to be unbounded and inferred in a data-driven fashion. Its construction induces sparsity, while at the same time allowing for more features to emerge, as new observations appear. Here, we focus on the stick-breaking construction of the IBP proposed by \cite{teh2007stick}, which renders it amenable to Variational Inference. Let us consider $N$ observations, and a binary matrix $\boldsymbol Z = [z_{j,k}]_{j,k=1}^{N,K}$; each entry therein, indicates the existence of feature $k$ in observation $j$. Taking the infinite limit $K\rightarrow \infty$, we can construct the following hierarchical representation \citep{teh2007stick, theo}:
	\begin{align*}
		\label{ibp_equations}
		u_k \sim \mathrm{Beta}(\alpha,1), \ \pi_k = \prod_{i=1}^k u_i, \ z_{j,k} \sim \mathrm{Bernoulli}(\pi_k), \ \forall j 
	\end{align*}
	where $\alpha$ is a non-negative parameter, controlling the induced sparsity.
	
	\subsection{Local Winner-Takes-All}
	
	Let us assume a single layer of an LWTA-based network, comprising $K$ LWTA blocks with $U$ competing units therein. Each block produces an output $\boldsymbol{y}_k \in \mathrm{one\_hot(U)}$, $k=1, \dots, K$, given some input $\boldsymbol x \in \mathbb{R}^{J}$. Each \textit{linear} unit in each block computes its activation $h_k^u, \ u=1, \dots U$, and the output of the block is decided via competition among its units. Thus, for each block, $k$, and unit, $u$, therein, the output reads:
	\begin{align}
		y_k^u = g(h_k^1, \dots, h_k^U)
	\end{align}
	where $g(\cdot)$ is the \textit{competition function}. The activation of each individual unit follows the conventional inner product computation $h_k^u = \boldsymbol{w}_{ku}^T \boldsymbol{x}$, where $\boldsymbol{W} \in \mathbb{R}^{J \times K \times U}$ is the weight matrix of the LWTA layer. In a conventional \emph{hard} LWTA network, the final output reads:
	\begin{align}
		y_k^u = 
		\begin{cases}
			1, & \text{if } h_k^u \geq h_k^i, \qquad \forall i=1, \dots, U, \ i \neq u\\
			0, & \text{otherwise}
		\end{cases}
	\end{align}
	To bypass the restrictive assumption of binary output, more expressive versions of the competition function have been proposed in the literature, e.g., \cite{srivastava2013compete}. These generalized \emph{hard} LWTA networks postulate:
	\begin{align}
		y_k^u = 
		\begin{cases}
			h_k^u, & \text{if } h_k^u \geq h_k^i, \qquad \forall i=1, \dots, U, \ i \neq u\\
			0, & \text{otherwise}
		\end{cases}
		\label{eq:deterministic_lwta}
	\end{align}
	This way, only the unit with the \textit{strongest} activation produces an output in each block, while the others are inhibited to silence, i.e., the zero value. This way, the output of each layer of the network yields a sparse representation according to the competition outcome within each block. 
	
	The above schemes do not respect a major aspect that is predominant in biological systems, namely \emph{stochasticity}. We posit that this aspect may be crucial for endowing deep networks with adversarial robustness. To this end, we adopt a scheme similar to \cite{panousis2019nonparametric}. That work proposed a novel competitive random sampling procedure. We explain this scheme in detail in the following Section.

	
	\section{Model Definition}
	\label{proposed_model_section}
	
	In this work, we consider a principled way of designing deep neural networks that renders their inferred representations considerably more robust to adversarial attacks. To this end, we utilize a novel stochastic LWTA type of activation functions, and we combine it with appropriate sparsity-inducing arguments from nonparametric Bayesian statistics.
	
	Let us assume an input $ \boldsymbol{X} \in \mathbb{R}^{N\times J}$ with $N$ examples, comprising $J$ features each. In conventional deep architectures, each hidden layer comprises nonlinear units; the input is presented to the layer, which then computes an affine transformation via the inner product of the input with weights $\boldsymbol{W} \in \mathbb{R}^{J \times K}$, producing outputs $ \boldsymbol{Y} \in \mathbb{R}^{N \times K}$. The described computation for each example $n$ yields $\boldsymbol{y}_n = \sigma(\boldsymbol{W}^{T} \boldsymbol{x}_n + \boldsymbol{b}) \in \mathbb{R}^K, \ n=1, \dots, N$, where $\boldsymbol{b}\in \mathbb{R}^K$ is a bias factor and $\sigma(\cdot)$ is a non-linear activation function, e.g. ReLU. An architecture comprises intermediate and output layers of this type.
	
	Under the proposed stochastic LWTA-based modeling rationale, singular units are replaced by LWTA blocks, each containing a set of  
	\emph{competing linear} units. Thus, the layer input is now presented to each different block and each unit therein, via different weights. Letting $K$ be the number of LWTA blocks and $U$ the number of competing units in each block, the weights are now represented via a three-dimensional matrix $\boldsymbol W \in \mathbb{R}^{J \times K \times U}$.
	
	Drawing inspiration from \cite{panousis2019nonparametric}, we postulate that the local competition in each block is performed via a competitive random sampling procedure. The higher the output of a competing unit, the higher the probability of it being the winner. However, the winner is selected stochastically.
	
	In the following, we introduce a set of discrete latent vectors $\boldsymbol \xi_n \in \mathrm{one\_hot}(U)^K$, in order to encode the outcome of the local competition between the units in each LWTA block of a network layer. For each data input, $\boldsymbol{x}_n$, the non-zero entries in the aforementioned one-hot representation denotes the winning unit among the $U$ competitors in each of the $K$ blocks of the layer.
	
	To further enhance the stochasticity and regularization of the resulting model, we turn to the nonparametric Bayesian framework. Specifically, we introduce a matrix of latent variables $\boldsymbol{Z} \in \{0, 1\}^{J \times K }$, to explicitly regularize the model by inferring whether the model actually needs to use each connection in each layer. Each entry,  $z_{j,k}$,  therein is set to one, if the $j^{th}$ dimension of the input is presented to the $k^{th}$ block, otherwise $z_{j,k} = 0$. We impose the sparsity-inducing IBP prior over the latent variables $z$ and perform inference over them. Essentially, if all the connections leading to some block are set to $z_{j,k} = 0$, the block is effectively zeroed-out of the network. This way, we induce sparsity in the network architecture.
	
	On this basis, we now define the output of a layer of the considered model, $\boldsymbol{y}_n \in \mathbb{R}^{K\cdot U}$, as follows:
	\begin{align}
		[\boldsymbol{y}_n]_{ku} = [\boldsymbol \xi_n ]_{ku} \sum_{j=1}^J (w_{j,k,u} \cdot z_{j,k}) \cdot [ \boldsymbol{x}_n]_j \in \mathbb{R}
		\label{eqn:layer_output_ff}
	\end{align}
	To facilitate a competitive random sampling procedure in a data-driven fashion, the latent indicators $\boldsymbol \xi_n$ are drawn from a posterior Categorical distribution. The concept is that the higher the output of a linear competing unit, the higher the probability of it being the winner. We yield:

	\resizebox{.85\linewidth}{!}{
		\begin{minipage}{\linewidth}
			\begin{align}
				q([\boldsymbol \xi_n]_k ) = \mathrm{Discrete} \left([\boldsymbol \xi_n]_k \Big | \mathrm{softmax}\left( \sum_{j=1}^J [w_{j,k,u}]_{u=1}^U \cdot z_{j,k} \cdot [\boldsymbol x_n]_j \right) \right)
			\end{align}
		\end{minipage}
	}

	Further, we postulate that the latent variables $z$ are drawn from Bernoulli posteriors, such that:
	\begin{align}
		q(z_{j,k}) = \mathrm{Bernoulli}(z_{j,k} | \tilde{\pi}_{j,k})
	\end{align}
	These are trained by means of variational Bayes, as we describe next, while we resort to fixed-point estimation for the weight matrices $\boldsymbol W$. 
	
	For the output layer of our approach, we perform the standard inner product computation followed by a softmax, while imposing an IBP over the connections, similar to the inner layers. 
	Specifically, let us assume a $C$-unit penultimate layer with input $\boldsymbol X \in \mathbb{R}^{N \times J}$ and weights $\boldsymbol{W} \in \mathbb{R}^{J \times C}$. We introduce an auxiliary matrix of latent variables $\boldsymbol{Z} \in \{ 0, 1\}^{J \times C}$. Then, the output $\boldsymbol{Y}\in \mathbb{R}^{N \times C} $ yields:
	\begin{align}
		y_{n,c}=\mathrm{softmax}\big(\sum_{j=1}^{J}\left(w_{j, c} \cdot z_{j, c}\right) \cdot\left[\boldsymbol{x}_{n}\right]_{j}\big) \in \mathbb{R}
	\end{align}
	where the latent variables in $\boldsymbol{Z}$ are drawn independently from a Bernoulli posterior:
	\begin{equation}\label{Bernouli_posteriors}
		q\left(z_{j, c}\right)=\operatorname{Bernoulli}\left(z_{j, c} | \tilde{\pi}_{j, c}\right)
	\end{equation}

	To perform variational inference for the model latent variables, we impose a symmetric Discrete prior over the latent indicators, $[\boldsymbol \xi_n]_k \sim \mathrm{Discrete}(1/U)$. On the other hand, the prior imposed over $\boldsymbol{Z}$ follows the stick-breaking construction of the IBP, to facilitate data-driven sparsity induction.
	
	The formulation of the proposed modeling approach is now complete. A graphical illustration of the resulting model is depicted in Fig. \ref{synopsis:wta}.
	
	\subsection{Convolutional Layers}
	
	Further, to accommodate architectures comprising convolutional operations, we devise a convolutional variant inspired from \cite{panousis2019nonparametric}. Specifically, let us assume input tensors $\{ \boldsymbol{X}_n\}_{n=1}^N \in \mathbb{R}^{H \times L \times C}$ at a specific layer, where $H, L, C$ are the height, length and channels of the input. We define a set of kernels, each with weights $\boldsymbol{W}_k \in \mathbb{R}^{h \times l \times C \times U}$, where $h,l,C, U$ are the kernel height, length, channels and \textit{competing feature maps}, and $k=1, \dots K$. Thus, analogously to the grouping of linear units in the dense layers, in this case, local competition is performed among \textit{feature maps}. \textit{Each kernel is treated as an LWTA block with competing feature maps}; each layer comprises multiple kernels. 
	
	We additionally consider analogous auxiliary binary latent variables $\boldsymbol{z}\in \{0,1\}^K$ to further regularize the convolutional layers. Here, we retain or omit full LWTA blocks (convolutional kernels), as opposed to single connections. This way, at a given layer of the proposed convolutional variant, the output $\boldsymbol Y_n \in \mathbb{R}^{H \times L \times K \cdot U}$ is obtained via concatenation along the last dimension of the subtensors:
	\begin{align}
		[\boldsymbol Y_n]_k = [ \boldsymbol \xi_n]_k \left( \boldsymbol (z_k \cdot W_k) \star \boldsymbol X_n \right) \in \mathbb{R} ^ {H \times L \times U}
	\end{align}
	where $\boldsymbol X_n$ is the input tensor for the $n^{th}$ data point, and ``$\star$'' denotes the convolution operation.  Turning to the competition function, we follow the same rationale, such that the sampling procedure is driven from the outputs of the competing feature maps:
	\resizebox{\linewidth}{!}{
		\begin{minipage}{\linewidth}
			\begin{align*}
				q([\boldsymbol{\xi}_n]_k) = \mathrm{Discrete}\Big ([\boldsymbol \xi_n]_k \Big | \mathrm{softmax}(\sum_{h',l'} [ \boldsymbol (z_k \cdot W_k) \star \boldsymbol X_n]_{h',l',u}\Big)
			\end{align*}
		\end{minipage}
	}

\begin{figure*}
		\begin{subfigure}[t]{.45\textwidth}
			\centering
			\resizebox{1.1\linewidth}{!}{\def\layersep{3.5cm}
\def\inputsize{2}
\def\wtablocks{2}
\def\neuronsep{5}
\def\outputsize{2}
\def\wtasep{2.5}
\def\wtablocks{3}
\def\unitsperblock{2}
\def\prob{0.55}

\begin{tikzpicture}[-,draw, node distance=\layersep, label/.style args={#1#2}{%
    postaction={ decorate,
    decoration={ markings, mark=at position #1 with \node #2;}}}]
    \tikzstyle{every pin edge}=[<-,shorten <=1pt]
    \tikzstyle{neuron}=[circle,draw,minimum size=12pt,inner sep=0pt]
    \tikzstyle{wta} = [rounded corners,rectangle, draw, minimum_height=3cm, minimum_width=2cm]

    \tikzstyle{input neuron}=[neuron, fill=white!50];
    \tikzstyle{output neuron}=[neuron, fill=white];
    \tikzstyle{hidden neuron}=[neuron, fill=white!50];
    \tikzstyle{hidden neuron activated}=[neuron, fill=white!50, very thick];
    \tikzstyle{wta block} =[wta];
    \tikzstyle{annot} = [text width=10em, text centered]

    \node[input neuron, pin=left:$x_1$] (I-1) at (0,-\neuronsep+\wtasep) {};
    \node[input neuron, pin=left:$x_J$] (I-2) at (0,-2*\neuronsep+\wtasep) {};
		
     	\matrix (H-1) at (\layersep, -\wtasep-0.*\wtasep) [row sep=4mm, column sep=2mm, inner sep=3mm, box, matrix of nodes] 
      {
        		\node[hidden neuron activated](o1-1){}; \\
        		\node[hidden neuron](o1-2){}; \\
	  };

	  
	  \matrix (H-2) at (\layersep, -3*\wtasep-0.*\wtasep) [row sep=4mm, column sep=2mm, inner sep=3mm, box, matrix of nodes] 
      {
        		\node[hidden neuron activated](o2-1){}; \\
        		\node[hidden neuron](o2-2){}; \\
	  };

	\matrix (H2-1) at (2*\layersep, -\wtasep-0.*\wtasep) [row sep=4mm, column sep=2mm, inner sep=3mm, box, matrix of nodes] 
      {
        		\node[hidden neuron](o21-1){}; \\
        		\node[hidden neuron activated](o21-2){}; \\
	  };

	  \matrix (H2-2) at (2*\layersep, -3*\wtasep-0.*\wtasep) [row sep=4mm, column sep=2mm, inner sep=3mm, box, matrix of nodes] 
      {
        		\node[hidden neuron](o22-1){}; \\
        		\node[hidden neuron activated](o22-2){}; \\
	  };
	  
        
    \foreach \name / \y in {1,...,\outputsize}
    		\node[output neuron] (O-\name) at (3*\layersep,-\neuronsep*\y+\wtasep) {};

    \path (I-1) -- (I-2) node [black, font=\Huge, midway, sloped] {$\dots$};
 	\path[black!100, line width=0.55pt, thick] (I-1.east) edge node[ above, rotate=5, black!100] {\scriptsize $z_{1,1}=1$} (o1-1);
    \path[black!100, line width=0.55pt, thick] (I-1.east) edge node[ below, rotate=-15, black!100] {\scriptsize $z_{1,1}=1$} (o1-2.west);
    
    \path[black!100, line width=0.55pt, thick] (I-1.east) edge (o2-1.west);
    \path[black!100, line width=0.55pt, thick] (I-1.east) edge (o2-2.west);

	\path[black!100, line width=0.55pt, thick] (I-2.east) edge (o1-1.west);
    \path[black!100, line width=0.55pt, thick] (I-2.east) edge (o1-2.west);
    
    \path[black!40] (I-2.east) edge node[above, rotate=5, black!100] {\scriptsize $z_{J,K}=0$} (o2-1.west);
    \path[black!40] (I-2.east) edge node[below, rotate=-15, black!100] {\scriptsize $z_{J,K}=0$} (o2-2.west);
    
    
            
    \path (H-1) -- (H-2) node [black, font=\Huge, midway, sloped] {$\dots$};
    \path (H2-1) -- (H2-2) node [black, font=\Huge, midway, sloped] {$\dots$};

    \path[color=black!100, line width=0.55pt, thick] (o1-1.east) edge (o21-1.west); 
    \path[color=black!100, line width=0.55pt, thick] (o1-1.east) edge (o21-2.west);

    \path[color=black!40] (o1-1.east) edge (o22-1.west);  
    \path[color=black!40] (o1-1.east) edge (o22-2.west); 
    
    \path[black!40] (o1-2.east) edge (o21-1.west);  
    \path[black!40] (o1-2.east) edge (o21-2.west); 

    
    \path[black!40] (o1-2.east) edge (o22-1.west);  
    \path[black!40] (o1-2.east) edge (o22-2.west);

    \path[color=black!100, line width=0.55pt, thick] (o2-1.east) edge (o21-1.west);  
    \path[color=black!100, line width=0.55pt, thick] (o2-1.east) edge (o21-2.west);
    
    \path[color=black!100, line width=0.55pt, thick] (o2-1.east) edge (o22-1.west);  
    \path[color=black!100, line width=0.55pt, thick] (o2-1.east) edge (o22-2.west);
    
    \path[black!40] (o2-2.east) edge (o21-1.west);  
    \path[black!40] (o2-2.east) edge (o21-2.west);
    
    \path[black!40] (o2-2.east) edge (o22-1.west);  
    \path[black!40] (o2-2.east) edge (o22-2.west);
    
    \path (O-1) -- (O-2) node [black, font=\Huge, midway, sloped] {$\dots$};

    
    
    

    
    
    

    \path[black!40] (o21-1.east) edge (O-1.west);  
    \path[black!40] (o21-1.east) edge (O-2.west);
    
    \path[black!100, line width=0.55pt, thick] (o21-2.east) edge (O-1.west);  
    \path[black!100, line width=0.55pt, thick] (o21-2.east) edge (O-2.west);

    \path[black!40] (o22-1.east) edge (O-1.west);  
    \path[black!40] (o22-1.east) edge (O-2.west);
    
    \path[black!100, line width=0.55pt, thick] (o22-2.east) edge (O-1.west);  
    \path[black!100, line width=0.55pt, thick] (o22-2.east) edge (O-2.west);

    

    \node[annot,above= -1mm of o1-1] (k) {\scriptsize $\xi$= 1};
    \node[annot,below= -1mm of o1-2] (k) {\scriptsize $\xi$= 0};
    
    \node[annot,above= -1mm of o22-1] (k) {\scriptsize $\xi$= 0};
    \node[annot,below= -1mm of o22-2] (k) {\scriptsize $\xi$= 1};
    \node[annot,above of=H-1, node distance=1.8cm] (hl) {SB-LWTA layer};
    \node[annot,above left=-3mm and -1.7cm of H-1] (k) {\scriptsize $1$};
    
    \node[annot,above left=-3mm and -1.7cm of H2-1] (k) {\scriptsize $1$};
    
    \node[annot,below left=-3mm and -1.7cm of H-2] (k) {\scriptsize $K$};
    
    \node[annot,below left=-3mm and -1.7cm of H2-2] (k) {\scriptsize $K$};
    
    \node[annot,above of=H2-1, node distance=1.8cm] (hl2) {SB-LWTA layer};
    \node[annot,left of=hl] {Input layer};
    \node[annot,right of=hl2] {Output layer};
\end{tikzpicture}
}
			\caption{}
			\label{synopsis:wta}
		\end{subfigure}
		\begin{subfigure}[t]{.55\textwidth}
			\centering
			\resizebox{.95\linewidth}{!}{\input{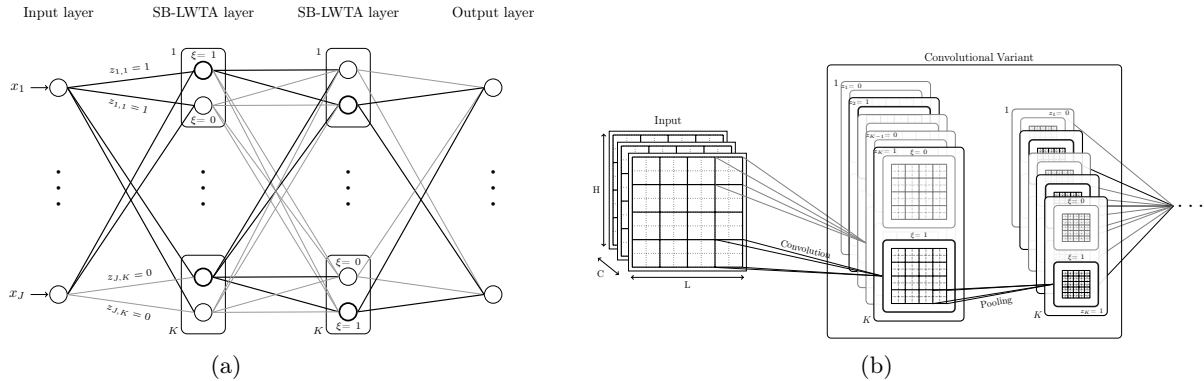}}
			\caption{}
			\label{synopsis:fig:cnn_sb_lwta}
		\end{subfigure}
		\caption{ (a) A graphical representation of our competition-based modeling approach. Rectangles denote LWTA blocks, and circles the competing units therein. The winner units are denoted with bold contours ($\xi=1$). Bold edges denote retained connections ($z=1$). (b) The convolutional LWTA variant. Competition takes place among feature maps. The winner feature map (denoted with bold contour) passes its output to the next layer, while the rest are zeroed out. }
	\end{figure*}

	We impose an IBP prior over $\boldsymbol{z}$, while a posteriori drawing from a Bernoulli distribution, such that,  $q(z_k) = \mathrm{Bernoulli}(z_k | \tilde{\pi}_k)$. We impose a symmetric prior for the latent winner indicators $ [\boldsymbol{\xi}_n]_k \sim \mathrm{Discrete}(1/U)$. A graphical illustration of the defined layer is depicted in Fig. \ref{synopsis:fig:cnn_sb_lwta}.
	
	\subsection{Training \& Inference}
	To train the proposed model, we resort to maximization of the Evidence Lower Bound (ELBO). To facilitate efficiency in the resulting procedures, we adopt Stochastic Gradient Variational Bayes (SGVB) \citep{kingma2014autoencoding}. However, our model comprises latent variables that are not readily amenable to the reparameterization trick of SGVB, namely, the discrete latent variables $z$ and $\boldsymbol \xi$, and the Beta-distributed stick variables $\boldsymbol u$. For the former, we utilize the continuous relaxation of Discrete (Bernoulli) random variables based on the Gumbel-Softmax trick \citep{maddison2016concrete,jang}. For the latter, we employ the Kumaraswamy distribution-based reparameterization trick \citep{kumaraswamy} of the Beta distribution. These reparameterization tricks are only employed during training to ensure low-variance ELBO gradients. 
	
	At inference time, we \emph{directly draw samples} from the trained posteriors of the \emph{winner and network subpart selection latent variables}  $\boldsymbol{\xi}$ and $z$, respectively; this introduces \emph{stochasticity to the network activations and architecture}, respectively. Thus, differently from previous work in the field, the arising stochasticity of the resulting model stems from two different sampling processes. On the one hand, contrary to deterministic competition-based networks presented in \cite{srivastava2013compete}, we implement a data-driven random sampling procedure to determine the winning units, by sampling from $q(\boldsymbol{\xi})$. In addition, we infer which subparts of the model must be used or omitted, again based on sampling from the trained posteriors $q(z)$\footnote{In detail, inference is performed by sampling the $q(\boldsymbol{\xi})$ and $q(z)$ posteriors a total of $S=5$ times, and averaging the corresponding $S=5$ sets of output logits. We have found that considering an increased $S>5$ does not yield any further improvement.}.    
	
	
	\section{Experimental Evaluation}
	
	We evaluate the capacity of our proposed approach against various adversarial attacks and under different setups. To this end, we follow the experimental framework of \cite{verma2019error}. Specifically, we train networks which either employ the standard one-hot representation to encode the output variable of the classifier (predicted class) or the error-correcting output strategy proposed in \cite{verma2019error}; the latter is based on Hadamard coding. We try various activation functions for the classification layer of the networks, and use either a single network or an ensemble, as described in \cite{verma2019error}. The details of the considered networks are provided in Table \ref{tab:models}. 
	To obtain some comparative results, the considered networks are formulated under both our novel deep learning framework and in a conventional manner (i.e., with ReLU nonlinearities and SGD training).

	\begin{table}[h]
		\centering
		\caption{A summary of the considered networks. $\mathbf{I}_k$ denotes a $k
			\times k$ identity matrix, while $\boldsymbol{H}_k$ a $k$-length Hadamard code.}
		\label{tab:models}
		\resizebox{\linewidth}{!}{
			\renewcommand{\arraystretch}{1.1}
			\begin{tabular}{lccc}
				\toprule
				Model & Architecture & Code & Output Activation\\\bottomrule
				Softmax & Standard &  $\mathbf{I}_{10}$ & softmax\\
				Logistic & Standard & $\mathbf{I}_{10}$ & logistic\\
				LogisticEns10 & Ensemble & $\mathbf{I}_{10}$ & logistic\\
				Tanh16 & Standard & $\mathbf{H}_{16}$ & tanh\\
				\bottomrule
			\end{tabular}
		}
	\end{table}

	\begin{table}
		\centering
		\caption{Classification accuracy on MNIST. }
		\label{MNIST_results}
		\resizebox{\linewidth}{!}{
			\begin{tabular}{ccccccc}
				\toprule 
				\addlinespace
				Model  & Params  & Benign  & PGD  & CW  & BSA  & Rand \tabularnewline
				\addlinespace
				\midrule 
				Softmax (U=4)  & 327,380  & .9613  & .865  & .970  & .950  & .187 \tabularnewline
				Softmax (U=2)  & 327,380  & .992  & \textbf{.935}  & .990  & \textbf{1.0}  & .961 \tabularnewline
				Logistic (U=2)  & 327,380  & .991  & .901  & .990  & .990  & .911 \tabularnewline
				LogEns10 (U=2)  & \textbf{205,190}  & .993  & .889  & .980  & .970  & \textbf{1.0}\tabularnewline
				\midrule 
				Madry  & 3,274,634  & .9853  & .925  & .84  & .520  & .351 \tabularnewline
				TanhEns16  & 401,168  & \textbf{.9948}  & .929  & \textbf{1.0}  & \textbf{1.0}  & .988 \tabularnewline
				\bottomrule
		\end{tabular}} 
	\end{table}

	\subsection{Experimental Setup}
	
	We consider two popular benchmark datasets, namely MNIST \citep{lecun2010mnist} and  CIFAR-10 \citep{Krizhevsky09learningmultiple}. The details of the considered network architectures are provided in the Supplementary.
	
	To evaluate our modeling approach, all the considered networks, depicted in Table \ref{tab:models}, are evaluated by splitting the architecture into: (i) LWTA blocks with $U=2$ competing units on each hidden layer, and (ii) blocks with 4 competing units. The total number of units on each layer (split into blocks of $U=2$ or 4 units) remains the same. 
	
	We initialize the posterior parameters of the Kumaraswamy distribution to $a=K, \ b=1$, where $K$ is the number of LWTA blocks, while using an uninformative Beta prior, $\text{Beta}(1,1)$. For the Concrete relaxations, we set the temperatures of the priors and posteriors to $0.5$ and $0.67$ respectively, as suggested in \cite{maddison2016concrete}; we faced no convergence issues with these selections.
	
	Evaluation is performed by utilizing 4 different benchmark adversarial attacks: (i) Projected Gradient Descent (PGD), (ii) Carlini and Wagner (CW), (iii) Blind Spot Attack (BSA) (\cite{zhang2019limitations}), and (iv) a random attack (Rand) \citep{verma2019error}. For all attacks, we adopt exactly the same experimental evaluation of \cite{verma2019error}, both for transparency and comparability. 
	
	Specifically, for the PGD attack, we use a common choice for the pixel-wise distortion $\epsilon=0.3 (0.031)$ for MNIST(CIFAR-10) with 500(200) attack iterations. For the CW attack, the learning rate was set to $1e$-$3$, utilizing 10 binary search steps. BSA performs the CW attack with a scaled version of the input, $\alpha \boldsymbol{x}$; we set $\alpha = 0.8$. In the ``Random'' attack, we construct random inputs by independently and uniformly generating pixels in $(0,1)$; we report the fraction of which that yield class probability less than $0.9$, in order to assess the confidence of the classifier as suggested in \cite{verma2019error}. 
	
	\subsection{Experimental Results}
	
	\textbf{MNIST}. We train our model for a maximum of 100 epochs, using the same data augmentation as in \cite{verma2019error}. In Table \ref{MNIST_results}, we depict the comparative results for each different network and adversarial attack. Therein, we also compare to the best-performing models in \cite{verma2019error}, namely the Madry model \citep{madry2017towards}, and TanhEns16. As we observe, our modeling approach yields considerable improvements over \cite{madry2017towards} and TanhEns16 \citep{verma2019error} in three of the four considered attacks, while imposing lower memory footprint (less trainable parameters). 
	
	
	\begin{table}[h!]
		\centering
		\caption{Classification accuracy on CIFAR-10. }
		\label{CIFAR_results}
		\resizebox{\linewidth}{!}{
			\begin{tabular}{ccccccc}
				\toprule 
				\addlinespace
				Model  & Params  & Benign  & PGD  & CW  & BSA  & Rand \tabularnewline
				\addlinespace
				\midrule
				Tanh16(U=4)  & 773,600  & .510  & .460  & .550  & .600  & .368 \tabularnewline
				Softmax(U=2)  & \textbf{772,628}  & .869  & .814  & \textbf{.860}  & \textbf{.870}  & .652 \tabularnewline
				Tanh16(U=2)  & 773,600  & .872  & \textbf{.826}  & .830  & \textbf{.830}  & .765 \tabularnewline
				LogEns10(U=2)  & 1,197,998  & .882  & .806  & .830  & .800  & \textbf{1.0} \tabularnewline
				\midrule 
				Madry  & 45,901,914  & .871  & .470  & .080  & 0  & .981 \tabularnewline
				TanhEns64  & 3,259,456  & \textbf{.896}  & .601  & .760  & .760  & \textbf{1.0} \tabularnewline
				\bottomrule
		\end{tabular}} 
	\end{table}
	
	\textbf{CIFAR-10.} For the CIFAR-10 dataset, we follow an analogous procedure. The obtained comparative effectiveness of our approach is depicted in Table \ref{CIFAR_results}. Therein, we also compare to the best-performing models in \cite{verma2019error}, namely the Madry model \citep{madry2017towards}, and TanhEns64. In this case, the differences in the computational burden are more evident, since now the trained networks are based on the VGG-like architecture. Specifically, our approach requires one-two orders of magnitude less parameters than the best performing alternatives in \cite{verma2019error}. At the same time, it yields substantially superior accuracy in the context of the considered attacks, while retaining a comparable performance for the benign case. 
	
	Note that the best performing model of \cite{verma2019error}, namely TanhEns64, adopts an architecture similar to Baseline Tanh, that we compare with, yet it fits an Ensemble of 4 different such networks.  This results in imposing almost 4 times the computational footprint of our approach, both in terms of memory requirements and computational times. This justifies the small improvement in accuracy in the benign case; yet, it renders our approach much more preferable in terms of the offered computation/accuracy trade-off.

	\begin{figure*}[ht!]
		\centering
		\begin{subfigure}[b]{0.24\textwidth}
			\includegraphics[width=\textwidth]{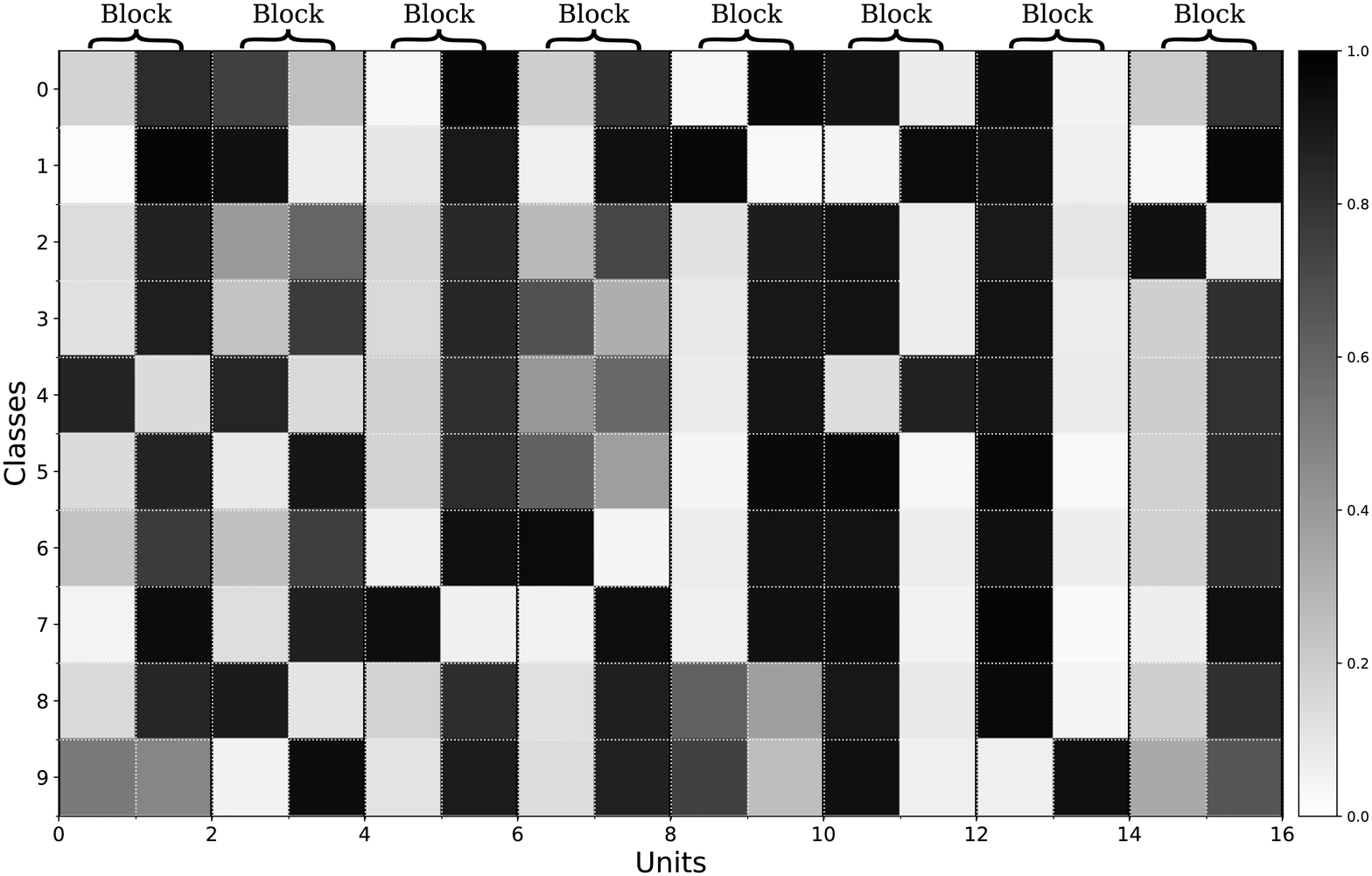}
			\caption{Benign: U=2}
			\label{probs_benign_U_2}
		\end{subfigure}
		\begin{subfigure}[b]{0.24\textwidth}
			\includegraphics[width=\textwidth]{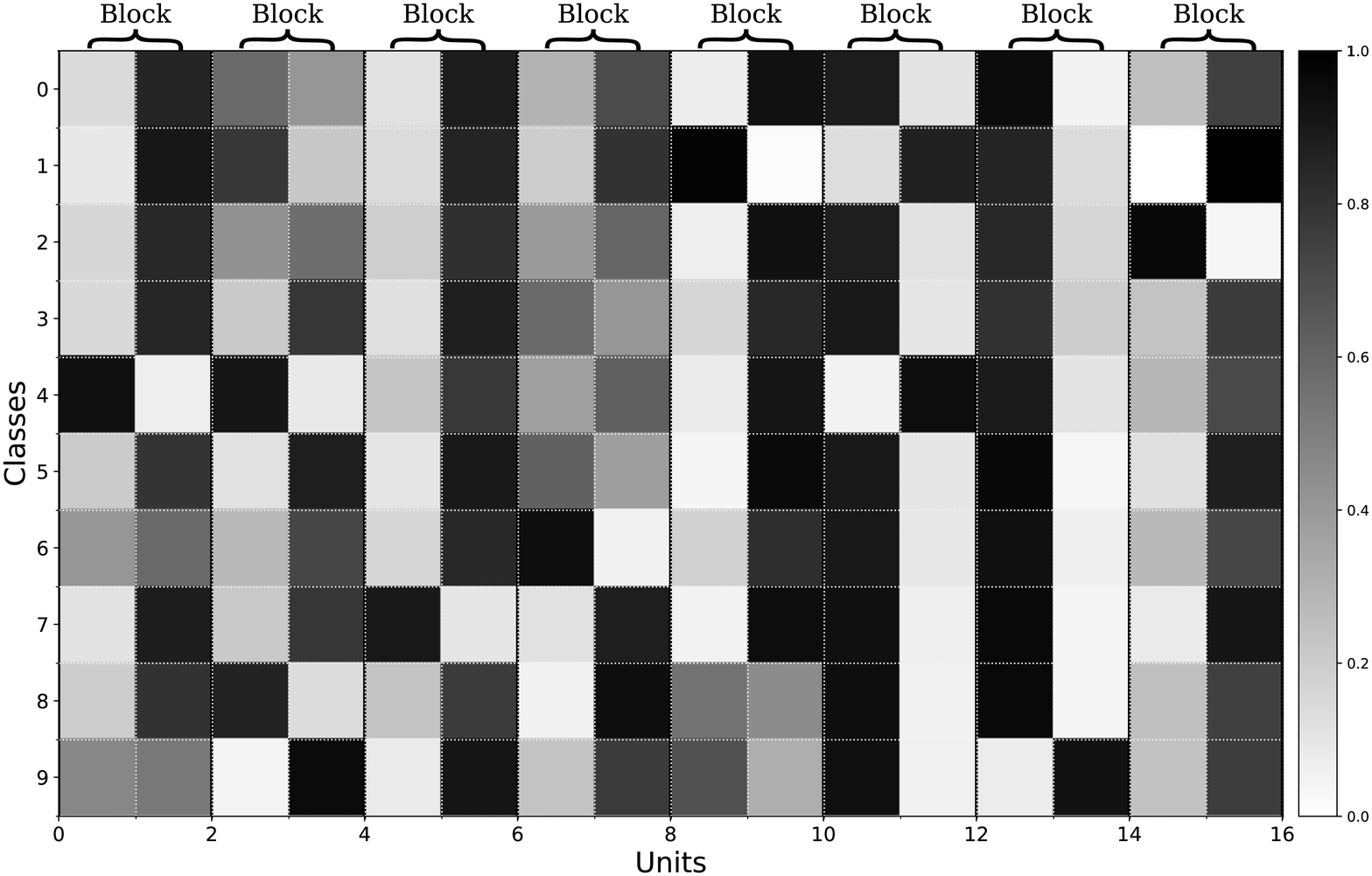}
			\caption{PGD: U=2}
			\label{probs_pgd_U_2}
		\end{subfigure}
		\begin{subfigure}[b]{0.25\textwidth}
			\includegraphics[width=\textwidth]{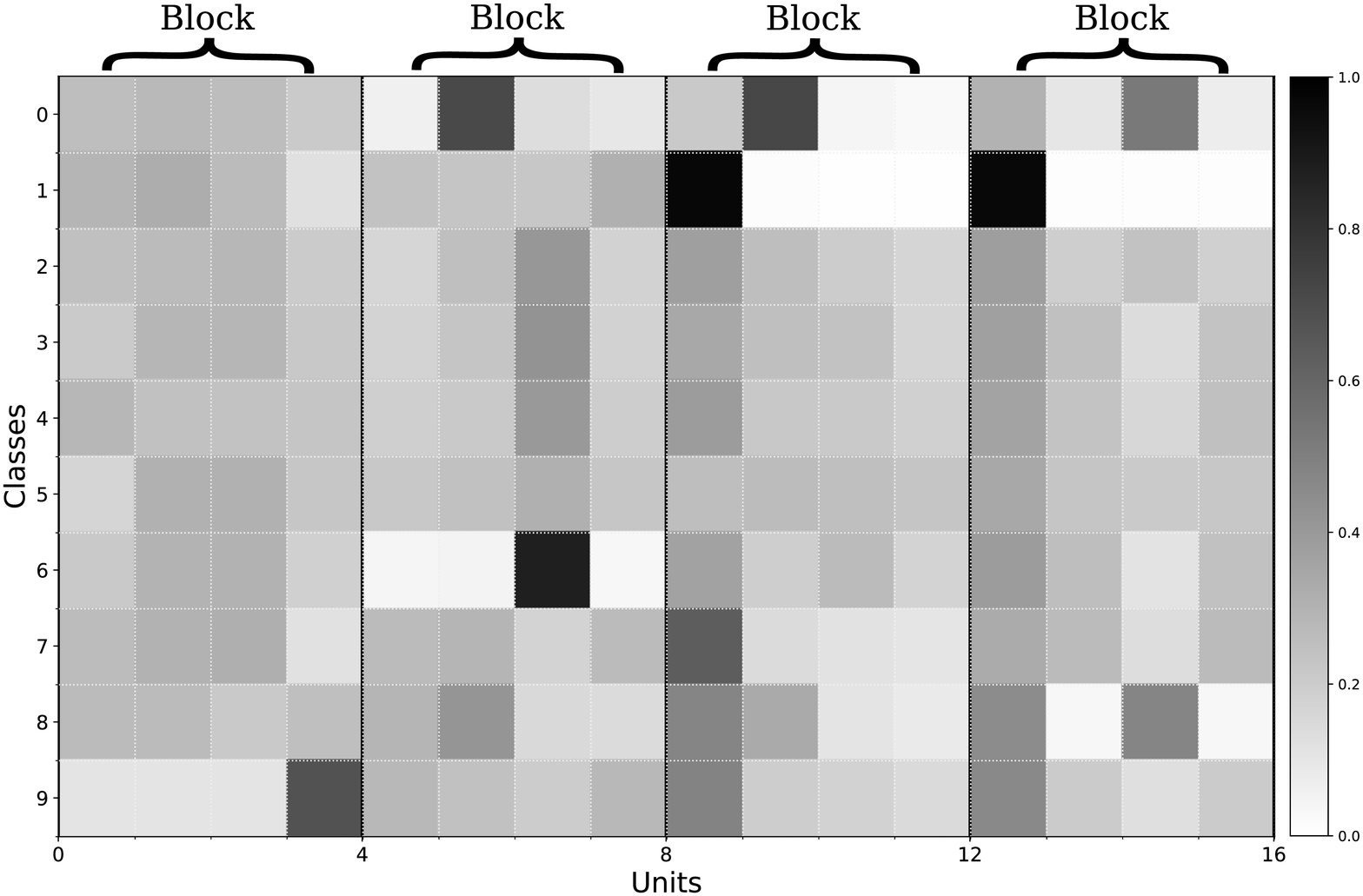}
			\caption{Benign: U=4}
			\label{probs_benign_U_4}
		\end{subfigure}
		\begin{subfigure}[b]{0.25\textwidth}
			\includegraphics[width=\textwidth]{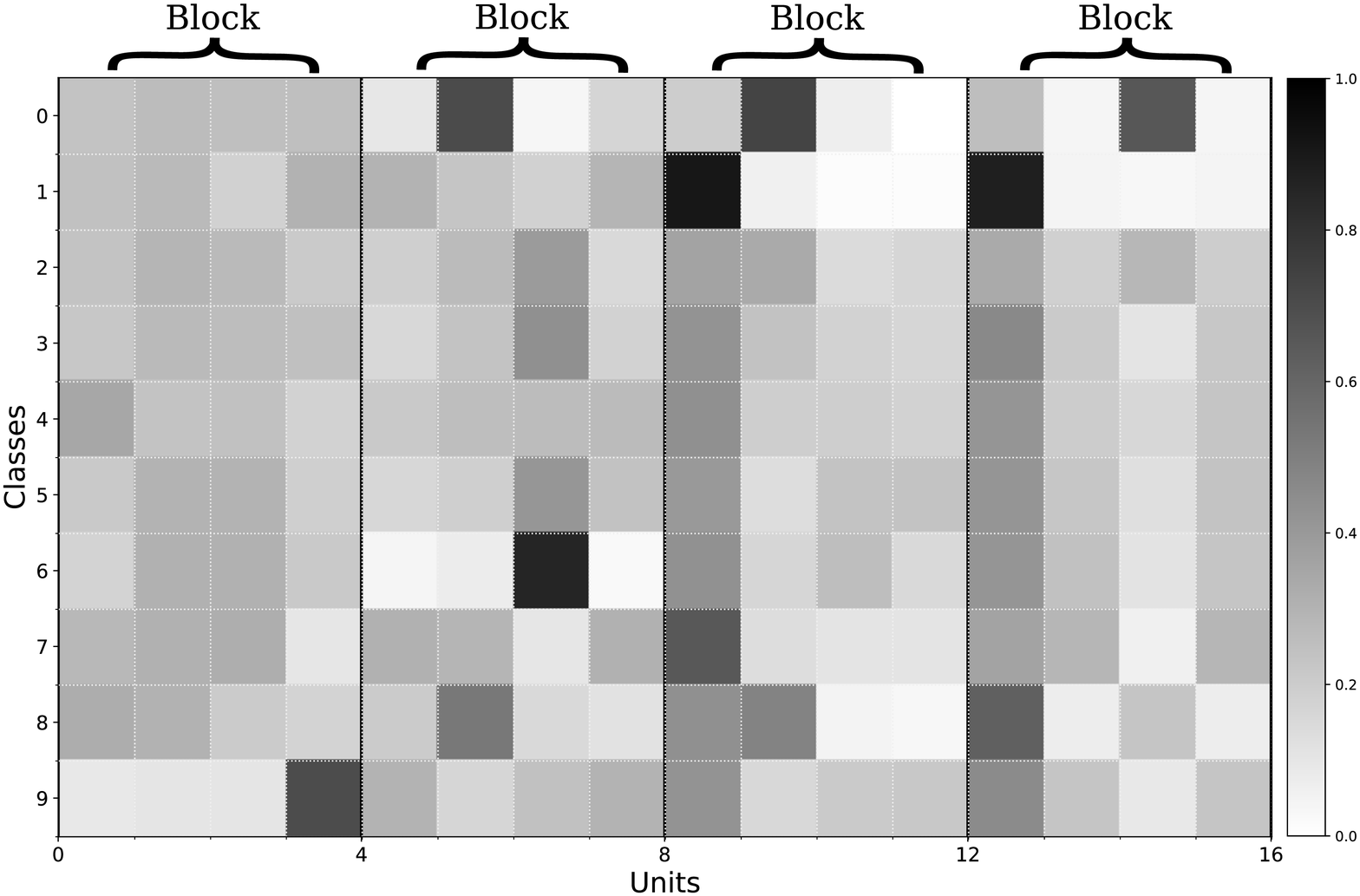}
			\caption{PGD: U=4}
			\label{probs_pgd_U_4}
		\end{subfigure}
		\caption{Winning probabilities of competing units in LWTA blocks, on an intermediate layer of the Tanh16 network, for each class in the CIFAR-10 dataset. Figs. \ref{probs_benign_U_2} and \ref{probs_pgd_U_2} depict the mean probability of activations per input class on a layer with 2 competing units, for benign and PGD test examples, respectively. Figs. \ref{probs_benign_U_4} and \ref{probs_pgd_U_4} correspond to a network layer comprising 4 competing units. Black denotes very high winning probability, while white very low.}
		\label{probabilities_figure}
	\end{figure*}
	
	\subsection{Further Insights}
	
	\begin{table}
		\centering
		\caption{Classification accuracy for all the attacks on the MNIST dataset (with $U=2$, wherever applicable).}
		\label{tab:mnist_ablation}
		\renewcommand{\arraystretch}{1.1}
		\resizebox{1.\linewidth}{!}{
			\begin{tabular}{cccccccc}
				\toprule
				Model & Method & Benign & PGD & CW & BSA & RAND\\\hline
				\multirow{5}{*}{Softmax}  & Baseline &  .992 & .082 & .540 & .180 & .270\\
				& LWTA\textsuperscript{max} & .993 & .302 & .800 & .390 & .543\\
				& LWTA\textsuperscript{max} \& IBP & \textbf{.994} & .890 & .920 & .840 & .361\\
				& LWTA & .992 & .900 & \textbf{.990} & \textbf{1.0} & .810\\
				& LWTA \& IBP & .992 & \textbf{.935} & \textbf{.990} & \textbf{1.0} & \textbf{.961}\\\bottomrule
				\multirow{5}{*}{Logistic} &  Baseline & .993 & .093 & .660 & .210 & .684\\
				& LWTA\textsuperscript{max} & .993 & .388 & .780 & .420 & .700\\
				& LWTA\textsuperscript{max} \& IBP & \textbf{.993} & .894 & .960 & .950 & .230\\
				& LWTA & .991 & .856 & \textbf{.990} & \textbf{.990} & \textbf{.982}\\
				& LWTA \& IBP & .991 & \textbf{.901} & \textbf{.990} & \textbf{.990} & .981\\\hline
				\multirow{5}{*}{LogisticEns10} & Baseline & .993 & .382 & .880 & .480 & .905\\ 
				& LWTA\textsuperscript{max} & .993 & .303 & .920 & .520 & .900\\
				& LWTA\textsuperscript{max} \& IBP& .\textbf{994} & .860  &  .940  & .910 & .400\\
				& LWTA & .992 & .603 & .900 & .550 & .809\\
				& LWTA \& IBP& .993 & \textbf{.889}  &  \textbf{.980}  & \textbf{.970} & \textbf{1.0}\\\hline
				\multirow{5}{*}{Tanh16}  & Baseline &  .993 & .421 & .790 & .320 & .673\\
				& LWTA\textsuperscript{max} & .992 & .462 & .910 & .420 & .573\\
				& LWTA\textsuperscript{max} \& IBP& .\textbf{994} & .898 & .960 & .940 & .363\\
				& LWTA & .992 & .862 & \textbf{.990} & \textbf{.980} & \textbf{.785}\\
				& LWTA \& IBP& .990 & \textbf{.900} &  \textbf{.990} & \textbf{.980} & \textbf{.785}\\\hline
			
			\end{tabular}
			
		}
	\end{table}
	
	\subsubsection{Ablation Study}
	Further, to assess the individual utility of LWTA-winner and architecture sampling in the context of our model, we scrutinize the obtained performance in both the benign case and the considered adversarial attacks. Specifically, we consider two different settings: (i) utilizing only the proposed LWTA mechanism in place of conventional ReLU activations; (ii) our full approach combining LWTA units with IBP-driven architecture sampling. The comparative results can be found in Tables \ref{tab:mnist_ablation} (MNIST) and \ref{tab:cifar_ablation} (CIFAR-10). Therein, "Baseline" corresponds to ReLU-based networks, as reported in \cite{verma2019error}.
	Our model's results are obtained with LWTA blocks of $U=2$ competing units. 
	
	We begin with the MNIST dataset, where we examine two additional setups. Specifically, we employ the deterministic LWTA competition function as defined in \eqref{eq:deterministic_lwta} and examine its effect against adversarial attacks, both as a standalone modification (denoted as LWTA\textsuperscript{max}) and also combined with the IBP-driven mechanism (i.e., sampling the $z$ variables). The resulting classification performance is illustrated in the second and third row for each network in Table \ref{tab:mnist_ablation}. One can observe that using deterministic LWTA activations, without stochastically sampling the winner, yields improvement, which increases with the incorporation of the IBP. 
	
	The fourth and fifth rows correspond to the \textit{proposed} LWTA (stochastic) activations, without or with architecture sampling (via the $z$ variables). The experimental results suggest that the stochastic nature of the proposed activation significantly contributes to the robustness of the model. It yields significant gains in all adversarial attacks compared to both the baseline, as well as to the deterministic LWTA adaptation. The performance improvement obtained by employing our proposed LWTA units can be as high as \emph{two orders of magnitude}, in the case of the powerful PGD attack. Finally, stochastic architecture sampling via the IBP-induced mechanism further increases the robustness.
	
	The corresponding experimental results for CIFAR-10 are provided in Table \ref{tab:cifar_ablation}. Our approach yields a significant performance increase, which reaches up to \emph{three orders of magnitude} (PGD attack, Logistic network).

	\begin{table}[h!]
		\centering
		\caption{Classification accuracy for all the attacks on CIFAR-10 (with $U=2$, wherever applicable).}
		\label{tab:cifar_ablation}
		\renewcommand{\arraystretch}{1.2}
		\resizebox{\linewidth}{!}{
			\begin{tabular}{ccccccc}
				\toprule 
				Model  & Method  & Benign  & PGD  & CW  & BSA  & RAND \tabularnewline
				\midrule 
				\multirow{3}{*}{Softmax} & Baseline  & .864  & .070  & .080  & .040  & .404 \tabularnewline
				& LWTA  & .867  & .804  & .820  & \textbf{.870}  & \textbf{.701} \tabularnewline
				& LWTA \& IBP  & \textbf{.869}  & \textbf{.814}  & \textbf{.860}  & \textbf{.870}  & .652 \tabularnewline
				\midrule 
				\multirow{3}{*}{Logistic}  & Baseline  & \textbf{.865}  & .006  & .140  & .100  & .492 \tabularnewline
				& LWTA  & .830  & .701  & .720  & .696  & .690 \tabularnewline
				& LWTA \& IBP  & .837  & \textbf{.738}  & \textbf{.800}  & \textbf{.820}  & \textbf{.726} \tabularnewline
				\midrule 
				\multirow{3}{*}{LogisticEns10}  & Baseline  & .877  & .100  & .240  & .140  & .495 \tabularnewline
				& LWTA  & .879  & .712  & .750  & .750  & .803 \tabularnewline
				& LWTA \& IBP  & \textbf{.882}  & \textbf{.806}  & \textbf{.830}  & \textbf{.800}  & \textbf{1.0} \tabularnewline
				\midrule 
				\multirow{3}{*}{Tanh16} & Baseline  & .866  & .099  & .080  & .100  & .700 \tabularnewline
				& LWTA  & .863  & .720  & .750  & .750  & \textbf{.800} \tabularnewline
				& LWTA \& IBP  & \textbf{.872}  & \textbf{.826}  & \textbf{.830}  & \textbf{.830}  & .765 \tabularnewline
				\midrule 
				&  &  &  &  &  & \tabularnewline
		\end{tabular}} 
	\end{table}

	\begin{figure*}[!ht]
		\centering
		\begin{subfigure}[b]{0.49\textwidth}
			\centering
			\includegraphics[width=0.8\textwidth]{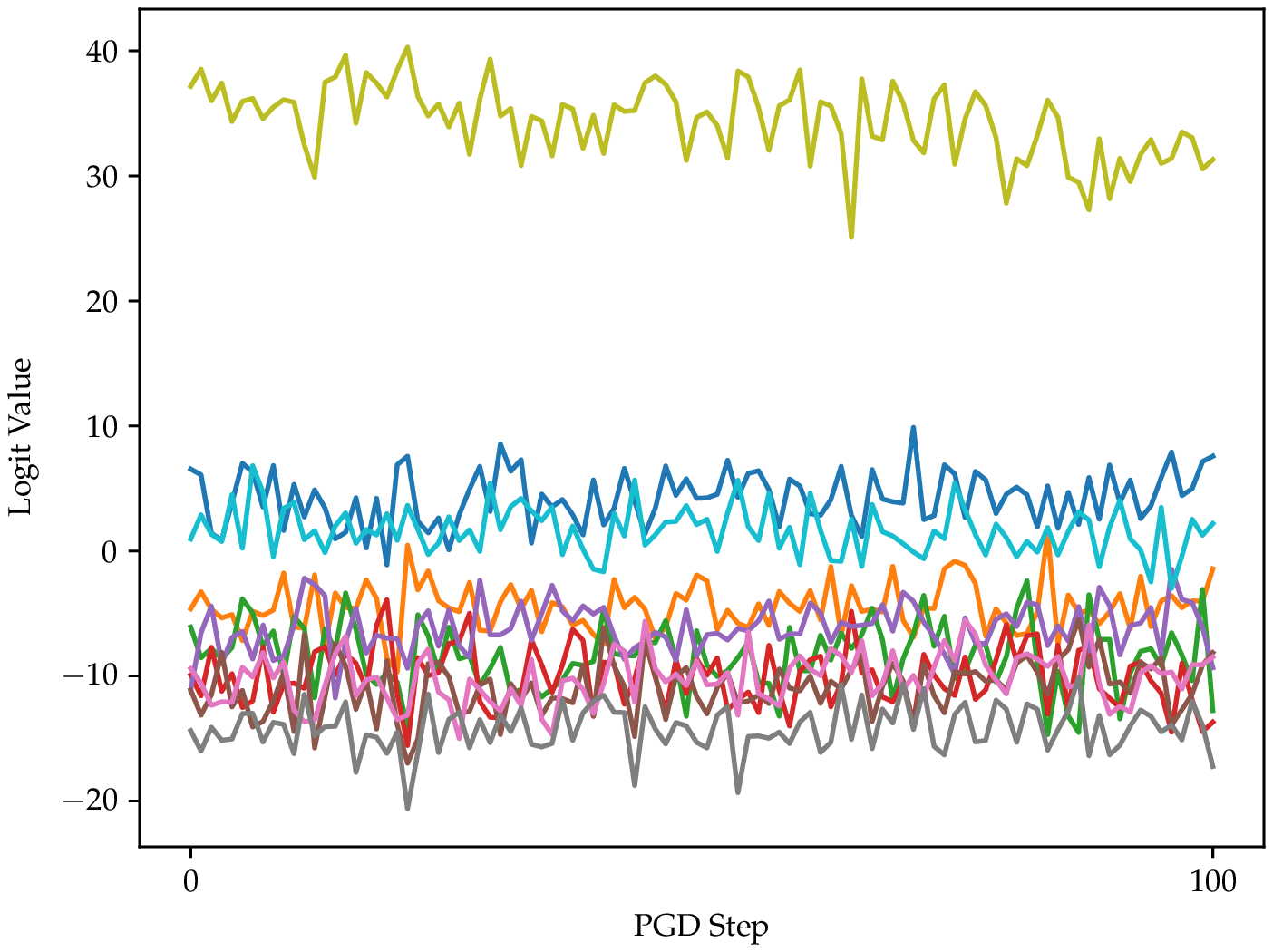}
			\caption{Proposed model}
			\label{lwta_logits}
		\end{subfigure}
		\begin{subfigure}[b]{0.49\textwidth}
			\centering
			\includegraphics[width=0.81\textwidth]{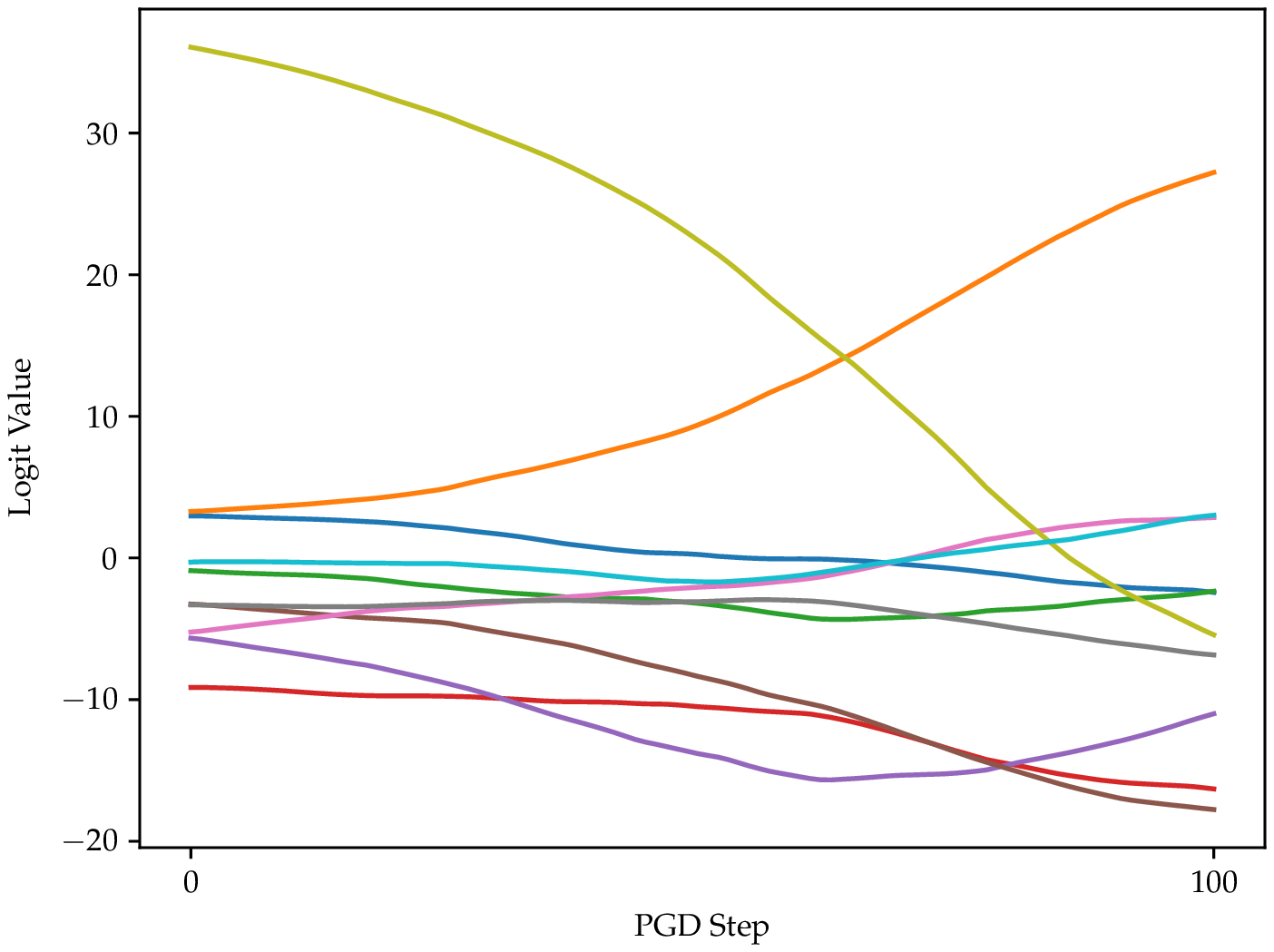}
			\caption{Conventional ReLU-based counterpart}
			\label{relu_logits}
		\end{subfigure}
		\caption{Change of the output logit values under our proposed approach (\ref{lwta_logits}), and a ReLU-based (\ref{relu_logits}) counterpart (PGD attack, CIFAR-10 dataset, Softmax network). The gradual, radical change of the logit values in the ReLU-based network indicates that conventional ReLU activations allow the attacker to successfully exploit gradient information. In contrast, under our proposed framework, the gradient-based attacker completely fails to do so. }
		\label{fig:logits_change}
	\end{figure*}

	\subsubsection{LWTA Behavior}

	
	Here, we scrutinize the competition patterns within the blocks of our model, in order to gain some further insights. First and foremost, we want to ensure that competition does not collapse to singular ``always-winning'' units. To this end, we choose a random intermediate layer of a Tanh16 network formulated under our modeling approach. We consider layers comprising 8 or 4 LWTA blocks of $U=2$ and 4 competing units, respectively, and focus on the CIFAR-10 dataset. The probabilities of unit activations for each class are depicted in Fig. \ref{probabilities_figure}.  
	
	In the case of $U=2$ competing units, we observe that the unit activation probabilities for each different setup (benign and PGD) are essentially the same (Figs. \ref{probs_benign_U_2} and \ref{probs_pgd_U_2}). This suggests that the LWTA mechanism succeeds in encoding salient discriminative patterns in the data that are resilient to PGD attacks. Thus, we obtain networks capable of defending against adversarial attacks in a principled way. 
	
	On the other hand, in Figs. \ref{probs_benign_U_4} and \ref{probs_pgd_U_4} we depict the corresponding probabilities when employing 4 competing units. In this case, the competition mechanism is uncertain about the winning unit in each block and for each class; average activation probability for each unit is around $\approx 25\%$. Moreover, there are several differences in the activations between the benign data and a PGD attack; these explain the significant drop in performance. This behavior potentially arises due to the relatively small structure of the network; from the 16 units of the original architecture, only 4 are active for each input. Thus, in this setting, LWTA fails to encode the necessary distinctive patterns of the data. Further results are provided in the Supplementary.
	
	Finally, we investigate how the classifier output logit values change in the context of an adversarial (PGD) attack. We expect that this investigation may shed more light to the distinctive properties of the proposed framework that allow for the observed adversarial robustness. To this end, we consider the Softmax network trained on the CIFAR-10 dataset; we use a single, randomly selected example from the test set to base our attack on. In Fig. \ref{fig:logits_change}, we depict how the logit values pertaining to the ten modeled classes vary as we stage the PGD attack on the proposed model, as well its conventional, ReLU-based counterpart. As we observe, the conventional, ReLU-based network exhibits a gradual, yet radical change in the logit values as the PGD attack progresses; this suggests that finding an adversarial example constitutes an ``easy'' task for the attacker. In stark contrast, our approach exhibits varying, inconsistent, and steadily minor logit value changes as the PGD attack unfolds; this is especially true for the dominant (correct) class. 
	
	This implies that, under our approach, the gradient-based attacker is completely obstructed from successfully exploiting gradient information. This non-smooth appearance of the logit outputs seems to be due to the doubly stochastic nature of our approach, stemming from LWTA winner ($\boldsymbol \xi$) and network component ($z$) sampling. Due to this stochasticity, different sampled parts of a network may contribute to the logit outputs each time. This destroys, with high probability, the linearity with respect to the input. Similar results on different randomly selected examples are provided in the Supplementary.

	\subsubsection{Effect on the Decision Boundaries}
	We now turn our focus to the classifier decision boundaries obtained under our novel paradigm. Many recent works \citep{fawziGeometrical, fawziEmpirical, ortizjimenez2020optimism} have shown that the decision boundaries of trained deep networks are usually highly sensitive to small perturbations of their training examples along some particular directions. These are exactly the directions that an adversary can exploit in order to stage a successful attack. In this context, \cite{OrtizModasHMT2020} showed that deep networks are especially susceptible to directions of discriminative features; they tend to learn low margins in their area, while exhibiting high invariance to other directions, where the obtained margins are substantially larger.

	\begin{figure}[h!]
		\centering
		\includegraphics[width = \columnwidth]{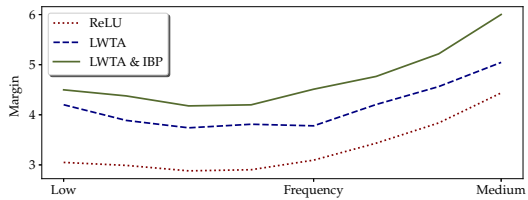}
		\caption{Mean margin over test examples of a LeNet-5 network trained on MNIST.}
		\label{fig:margins}
	\end{figure}

	First, we train a LeNet-5 network on the MNIST dataset, similar to \cite{OrtizModasHMT2020}. We, then, extract the mean margin of the decision boundary, by using a subspace-constrained version of DeepFool \citep{deepfool}. This measures the margin of $M$ samples on a sequence of subspaces; these are generated by using blocks from a 2-dimensional discrete cosine transform (2D-DCT) \citep{dct}. Our results are depicted in Fig. \ref{fig:margins}. We train the network using our full model (dubbed "LWTA \& IBP"), as well as a version keeping only the proposed LWTA activations. As we observe, our approach yields a significant increase of the margin in the low to medium frequencies, exactly where the baseline is adversarially brittle. It is also notable that the use of the proposed IBP-driven architecture sampling mechanism is complementary to the proposed (stochastic) LWTA-type activation functions. 
	
\begin{figure}[h!]
		\centering
		\includegraphics[width = \columnwidth]{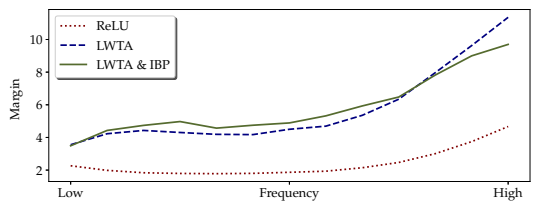}
		\caption{Mean margin over test examples on a VGG-like network trained on CIFAR-10.}
		\label{fig:margin_cifar}
	\end{figure}

	Subsequently, we repeat our experiments using a VGG-like architecture trained on CIFAR-10. We consider: (i) the standard, ReLU-based approach; (ii) our model using the proposed (stochastic) LWTA units but without IBP-driven architecture sampling; and (iii) the full model proposed in this work ("LWTA \& IBP"). The corresponding illustrations are depicted in Fig. \ref{fig:margin_cifar}. As we observe, by using the proposed stochastic LWTA-type units, we obtain a very significant increase of the margin across the frequency spectrum. Once again, the IBP-driven architecture sampling process presents further improvements to the network, especially at the lower end of the spectrum, where it is most needed.

	\begin{table}[ht!]
		\caption{Performance of a Softmax network \citep{verma2019error} under adversarial (FGSM-based) training for the MNIST dataset. Block size is $U=2$, wherever applicable.}
		\centering \resizebox{\linewidth}{!}{ %
			\begin{tabular}{cccccc}
				\toprule 
				\addlinespace
				Method  & Benign  & PGD  & CW  & BSA  & RAND \tabularnewline
				\midrule 
				\addlinespace
				Baseline (ReLU)  & \textbf{.992}  & .082  & .540  & .180  & .270 \tabularnewline
				LWTA  & \textbf{.992}  & .900  & \textbf{.990}  & \textbf{1.0}  & .810 \tabularnewline
				LWTA \& IBP  & \textbf{.992}  & \textbf{.935}  & \textbf{.990}  & \textbf{1.0}  & \textbf{.961} \tabularnewline
				\midrule 
				Baseline (ReLU) + FGSM  & \textbf{.992}  & .755  & .820  & .130  & .835 \tabularnewline
				LWTA + FGSM  & .991  & .925  & \textbf{.990}  & \textbf{1.0}  & .960 \tabularnewline
				LWTA \& IBP + FGSM  & .991  & \textbf{.970}  & \textbf{.990}  & \textbf{1.0}  & \textbf{.985} \tabularnewline
				\bottomrule
		\end{tabular}}
		\label{fgsm_results} 
	\end{table}

	\begin{table}
		\caption{Performance of a Softmax network \citep{verma2019error} under adversarial (FGSM-based) training for the CIFAR-10 dataset. Block size is $U=2$, wherever applicable.}
		\centering \resizebox{\linewidth}{!}{ %
			\begin{tabular}{cccccc}
				\toprule 
				\addlinespace
				Method  & Benign  & PGD  & CW  & BSA  & RAND \tabularnewline
				\midrule 
				\addlinespace
				Baseline (ReLU)  & .864  & .070  & .080  & .040  & .404 \tabularnewline
				LWTA  & .867  & .804  & .820  & \textbf{.870}  & \textbf{.701} \tabularnewline
				LWTA \& IBP  & \textbf{.869}  & \textbf{.814}  & \textbf{.860}  & \textbf{.870}  & .652
				\tabularnewline
				\midrule 
				Baseline (ReLU) + FGSM  & .780  & .200  & .380  & .340  & .185 \tabularnewline
				LWTA + FGSM  & .820  & .812  & .810  & \textbf{.870} & .440 \tabularnewline
				LWTA \& IBP + FGSM  & \textbf{.830}  & \textbf{.825}  & \textbf{.870}  & \textbf{.870}  & \textbf{.790} \tabularnewline
				\bottomrule
		\end{tabular}} 
		\label{fgsm_results_cifar} 
	\end{table}

	\subsection{Adversarial Training}
	Finally, it is important to analyze the behavior of our model under an adversarial training setup. Existing literature in the field has shown that conventionally-formulated networks lose some test-set performance when trained with adversarial examples. On the other hand, this sort of training renders them more robust when adversarially-attacked.
	
	Therefore, it is interesting to examine how networks formulated under our model perform in the context of an adversarial training setup. To this end, and due to space limitations, we limit ourselves to FGSM-based \citep{fgsm} adversarial training of the Softmax network. In the case of the MNIST dataset,  FGSM is run with $\epsilon=0.3$; for CIFAR-10, we set $\epsilon=0.031$. On this basis, we repeat the experiments of Section 4.3.1, and assess model performance considering all the attacks therein, under the same configuration; only exception to this rule is the PGD attack, where we use 40 and 20 PGD steps to attack the networks trained on MNIST and CIFAR-10, respectively. 
	
	We depict the obtained results in Tables \ref{fgsm_results} and \ref{fgsm_results_cifar}. Therein, the first three lines pertain to training by only making use of "clean" training data; the last three pertain to FGSM-based adversarial training, as described above. As we observe, in both datasets our (stochastic) LWTA activations completely outperform the ReLU-based baselines, while IBP-driven architecture sampling offers a further improvement in performance, across all attacks.

	\section{Conclusions}
	This work attacked adversarial robustness in deep learning. We introduced novel network design principles, founded upon stochastic units formulated under a sampling-based Local-Winner-Takes-All mechanism.
	We combined these with a network subpart omission mechanism driven from an IBP prior, which further enhances the stochasticity of the model. Our experimental evaluations have provided strong empirical evidence for the efficacy of our approach. Specifically, we yielded good accuracy improvements in various kinds of adversarial attacks, with considerably less trainable parameters. The obtained performance gains remained important under FGSM-based adversarial training. 
	
	Our future work targets novel methods for adversarial training deep networks formulated under our modeling approach. We specifically target scenarios that are not limited to the existing paradigm of gradient-based derivation of the adversarial training examples.
	
	\section*{Acknowledgments}
	This work has received funding from the European Union's Horizon 2020 research and innovation program under grant agreement No 872139, project aiD.

	\bibliography{main}
\end{document}